\documentclass[letterpaper]{article} 
\usepackage{aaai2026}  
\usepackage{times}  
\usepackage{helvet}  
\usepackage{courier}  
\usepackage[hyphens]{url}  
\usepackage{graphicx} 
\usepackage{natbib}  
\usepackage{caption} 
\usepackage{amsmath}
\usepackage{amssymb}
\usepackage{booktabs}
\usepackage{multirow}

\nocopyright 

\title{When Calibration Depends on the Scoring Rule:\\Quantized Biomedical LLM Classification}
\author{
    Anton Rasmussen\textsuperscript{1},
    Hong Qin\textsuperscript{1,2}
}
\affiliations{
    \textsuperscript{1}Department of Computer Science, Old Dominion University\\
    \textsuperscript{2}School of Data Science, Old Dominion University\\
    Norfolk, VA 23529, USA\\
    \{arasm003, hqin\}@odu.edu
}

\begin{document}

\maketitle

\begin{abstract}
Quantized large language models can run on consumer hardware, which motivates interest in on-premises processing of sensitive data.
Before such models can support evidence synthesis, however, their confidence estimates must be trustworthy.
Their reliability depends on implementation choices (prompt template, label wording, and scoring normalization) that are seldom treated as experimental variables.
We evaluate three 7-billion-parameter Mistral variants (a base model, BioMistral, and an instruction-tuned checkpoint evaluated without its chat template) at FP16, INT8, and INT4 on five-class sentence classification in medical abstracts.
We test two primary templates on $n = 2{,}000$ test sentences and two auxiliary templates on $n = 200$ validation sentences.
Our central observation, made without post-hoc calibration, is that switching from summed to mean-token log-likelihood reverses which model appears better calibrated: BioMistral's mean calibration error across matched conditions nearly triples, while the instruction-tuned model's drops by more than half.
Accuracy changes by at most 1.4 percentage points for these two checkpoints.
Negative log-likelihood and Brier score show the same reversal, so it is not specific to ECE binning.
The two primary templates were selected using test-derived examples, so absolute performance with them is exploratory.
Between them, prompt choice changes mean accuracy across precisions by 2.9 to 17.8 percentage points, depending on the model.
Eight-bit quantization changes accuracy by at most 1.1 percentage points for the adapted checkpoints; four-bit quantization shows mixed but non-catastrophic effects.
Post-hoc temperature scaling reduces calibration error under summed scoring but was not fitted under mean-token scoring, so whether the reversal survives per-scorer calibration is unknown.
These exploratory results suggest that calibration comparisons of decoder-based classifiers should treat scoring normalization and prompt design as first-order experimental decisions.
\end{abstract}

\section{Introduction}

Systematic reviews are central to evidence-based medicine, yet the manual screening and classification of research abstracts remains a major bottleneck.
Automated sentence classification, which assigns functional roles such as \textsc{Background}, \textsc{Objective}, \textsc{Methods}, \textsc{Results}, and \textsc{Conclusions} to sentences within medical abstracts, can accelerate evidence synthesis by structuring unstructured literature at scale~\citep{dernoncourt2017}.

Large language models (LLMs) are natural candidates for this task, and BioMistral-7B~\citep{labrak2024} exemplifies efforts to adapt general-purpose decoders to biomedical and medical NLP.
However, for evidence synthesis to be trustworthy, models must provide not only accurate classifications but also \emph{well-calibrated confidence estimates}.

Model compression makes local deployment feasible.
Techniques such as LLM.int8()~\citep{dettmers2022llmint8} roughly halve inference-time weight memory relative to FP16, making single-GPU inference practical for many 7-billion-parameter models.
Local deployment can keep model inputs within institutional infrastructure rather than sending them to third-party inference services.
Whether compression preserves \emph{calibration quality} is poorly understood, particularly for biomedical LLMs where the stakes of miscalibration are high.
Local deployment motivates our study but is not what we measure: all experiments use public PubMed abstracts, and we do not evaluate privacy protection or clinical deployment safety.

Instruction tuning and reinforcement learning from human feedback can improve instruction following, human preference, and selected behavioral measures~\citep{ouyang2022training}, but recent studies find that post-training and the chat templates that accompany it both degrade probability calibration~\citep{huang2026multilingual, sanzguerrero2026overconfident}.
Apparent calibration effects are also entangled with the probability-extraction protocol: the prompt template, verbalizer, scoring rule, and chat format all influence the induced class distribution.
Our exploratory results suggest that these measurement choices can dominate the apparent model comparison.

This paper makes three contributions within this exploratory design:

\begin{enumerate}
\item \textbf{Scoring-normalization sensitivity.} We observe that switching from summed to mean-token log-likelihood scoring reverses which of two checkpoints, BioMistral-7B and Mistral-7B-Instruct-v0.3, appears better calibrated on PubMed RCT before any post-hoc calibration, while accuracy remains largely unchanged.
This observation motivates specifying and varying the scoring protocol in calibration comparisons.

\item \textbf{Template variation.} Between the two primary templates, evaluated on the same test sample, mean accuracy over the three precisions differs by ${\sim}$3pp for Instruct, ${\sim}$7pp for BioMistral, and up to ${\sim}$18pp for the base model, and the template determines which model achieves the highest accuracy.

\item \textbf{Quantization deltas.} For the adapted checkpoints (BioMistral, Instruct), INT8 quantization changes accuracy by at most 1.1pp relative to FP16.
The base model shows larger INT8 effects on some templates (up to $+$4.2pp on Struct-Delim).
INT4 (NF4) shows heterogeneous effects that vary by model and template, without catastrophic degradation in the configurations tested.
\end{enumerate}

\section{Related Work}

\paragraph{Calibration of neural networks.}
\citet{guo2017} demonstrated that temperature scaling is a simple and effective post-hoc calibration method.
For pretrained transformers, calibration varies across tasks and distribution shifts, and temperature scaling can improve in-domain calibration~\citep{desai2020calibration}.
For RLHF-tuned LLMs, \citet{tian2023just} find that verbalized confidences are typically better calibrated than conditional token probabilities.

\paragraph{Instruction tuning and calibration.}
\citet{huang2026multilingual} show that instruction tuning can raise confidence without commensurate accuracy gains, worsening calibration across multiple languages.
\citet{sanzguerrero2026overconfident} decouple post-training from chat templates, finding that templates independently aggravate miscalibration via an ``ownership bias'' (up to 26\% higher confidence on self-generated answers).
These findings motivate our analysis: apparent calibration effects may reflect measurement artifacts.

\paragraph{Verbalizer sensitivity and surface form competition.}
When decoder LMs are used as classifiers, predicted probabilities depend on the \emph{verbalizer}, the mapping from task labels to model-facing output tokens.
\citet{holtzman2021surface} introduced the concept of surface form competition, showing that a model's probability mass is split among surface-equivalent continuations, so the highest-probability answer is not always the semantically correct one.
\citet{cho2026choices} demonstrate that raw and length-normalized likelihood scoring remain sensitive to superficial answer-choice content and input format, calling into question standard LLM evaluation protocols.
\citet{zhao2021calibrate} show that few-shot prompting introduces strong recency, majority-label, and common-token biases into LLM predictions, proposing content-free calibration to mitigate them.
These results bear on our work: the five PubMed RCT label names have unequal token lengths (3 to 5 tokens), creating a surface-form length prior that interacts with the scoring policy (summed vs.\ mean log-likelihood).

\paragraph{Biomedical LLMs.}
BioMistral-7B~\citep{labrak2024} continues pretraining of Mistral-7B-Instruct-v0.1 on PubMed Central articles, achieving competitive results on biomedical benchmarks.
PubMedBERT~\citep{gu2021pubmedbert} demonstrated that domain-specific pretraining from scratch outperforms general-domain models on biomedical NLP tasks.
\citet{labrak2024} include quantized variant evaluation and some calibration analysis, but do not vary the scoring rule or verbalizer.

\paragraph{Model quantization.}
LLM.int8()~\citep{dettmers2022llmint8} introduced mixed-precision INT8 decomposition that preserves outlier features in FP16, achieving near-lossless compression.
QLoRA~\citep{dettmers2023qlora} introduced NF4 quantization with double quantization for efficient fine-tuning.
While predictive performance under quantization has been studied~\citep{dettmers2022llmint8}, calibration effects have received less attention.
\citet{zhong2025quantized} study calibration in quantized LLMs across multiple models and datasets, finding that quantization generally worsens calibration and proposing a soft-prompt recovery framework.
Our work complements theirs by examining answer-text scoring sensitivity within the quantized biomedical setting.

\paragraph{Automated prompt optimization.}
EvoPrompting~\citep{chen2023evoprompting} used language models as mutation and crossover operators in evolutionary neural architecture search.
AlphaEvolve~\citep{alphaevolve2025} extends this paradigm to general program optimization, using LLM-proposed code changes and automated evaluation metrics.
We use AlphaEvolve with our own evaluator for a small prompt search in biomedical classification template design.

\section{Methods}

\subsection{Task and Dataset}

We evaluate on PubMed RCT sentence classification~\citep{dernoncourt2017}, a 5-class task assigning functional roles (\textsc{Background}, \textsc{Objective}, \textsc{Methods}, \textsc{Results}, \textsc{Conclusions}) to sentences from medical abstracts.
We use a stratified sample of $n = 2{,}000$ sentences from $1{,}362$ abstracts in the test set for the main evaluation and, for temperature scaling, $n = 200$ validation-set examples for fitting plus a separate $n = 200$ test sample for evaluation.
Paired confidence intervals use abstract-cluster bootstrap resampling ($5{,}000$ resamples) to account for within-abstract sentence correlation.

\subsection{Models}

\paragraph{Mistral-7B-v0.3}\footnote{\texttt{mistralai/Mistral-7B-v0.3}} is the 7B-parameter base decoder with neither task-specific fine-tuning nor instruction tuning.
It serves as the generic LLM baseline for comparison with the adapted checkpoints below.

\paragraph{BioMistral-7B}\footnote{\texttt{BioMistral/BioMistral-7B}}~\citep{labrak2024} was obtained by continuing the pretraining of Mistral-7B-Instruct-v0.1 on PubMed Central articles.
BioMistral inherits instruction tuning from its parent; the comparison with Mistral-Instruct-v0.3 therefore contrasts \emph{domain-continued} pretraining against a \emph{newer instruction-tuned} checkpoint, not a clean domain-vs-instruction contrast.

\paragraph{Mistral-7B-Instruct-v0.3}\footnote{\texttt{mistralai/Mistral-7B-Instruct-v0.3}} is the instruction-tuned variant of Mistral-7B-v0.3.
All three decoder models share the same 7B Mistral architecture but differ in training history and data.
We therefore treat them as specific checkpoints and evaluation configurations, and we do not attribute differences among them to domain adaptation or instruction tuning in isolation.
\textbf{Note:} We do not apply the Mistral chat template to the Instruct model; all models receive identical raw prompt text.
This choice isolates the template-and-scoring variables under study, but means that Instruct is evaluated outside its intended input format, which may independently affect calibration~\citep{sanzguerrero2026overconfident}.

\paragraph{PubMedBERT}\footnote{\texttt{microsoft/BiomedNLP-BioMedBERT-base-\allowbreak{}uncased-abstract-fulltext}}~\citep{gu2021pubmedbert} is a 110M-parameter encoder pretrained from scratch on PubMed abstracts and PubMed Central full text, then fine-tuned with a classification head.
It serves as a supervised encoder reference, evaluated on a separate balanced sample.

Table~\ref{tab:lineage} summarizes model provenance. We used the default Hugging Face Hub revisions available in August 2026; we did not record immutable checkpoint identifiers, which limits exact replication of checkpoint state.

\begin{table}[t]
\caption{Model lineage. BioMistral descends from Instruct-v0.1 (not v0.3); comparisons across rows do not isolate a single training variable. CPT denotes continued pretraining.}
\label{tab:lineage}
\centering
\scriptsize
\setlength{\tabcolsep}{3pt}
\begin{tabular}{@{}llll@{}}
\toprule
\textbf{Model} & \textbf{Parent} & \textbf{Training} & \textbf{Chat fmt} \\
\midrule
Base-v0.3   & --- & Pretrain only & None \\
Instruct-v0.3 & Base-v0.3 & + instruction tuning & Not applied \\
BioMistral  & Instruct-v0.1 & + PubMed CPT & Not applied \\
PubMedBERT  & from scratch & 176K supervised & N/A \\
\bottomrule
\end{tabular}
\end{table}

\subsection{Scoring Protocols}

\paragraph{Answer-text log-likelihood (decoders).}
For each example and candidate label, we compute the sum of the token log probabilities of the label text (e.g., ``BACKGROUND'') conditioned on the prompt:
\[
\text{score}(\ell) = \sum_{t=1}^{T_\ell} \log P(w_t^{(\ell)} \mid \text{prompt}, w_1^{(\ell)}, \ldots, w_{t-1}^{(\ell)}),
\]
where $w_1^{(\ell)}, \ldots, w_{T_\ell}^{(\ell)}$ are the tokens of label $\ell$.
Because label names have unequal token counts ($T_\ell = 3$ to $5$), we also evaluate a length-normalized variant:
\begin{equation*}
\text{score}_{\text{mean}}(\ell) = \frac{1}{T_\ell}\sum_{t=1}^{T_\ell} \log P\!\left(w_t^{(\ell)} \,\middle|\, \text{prompt}, w_{<t}^{(\ell)}\right).
\end{equation*}
We softmax-normalize the candidate scores (sum or mean) across the five labels to obtain a probability distribution.
The two scorers are not symmetric alternatives: the softmax over summed log-likelihoods is the model's conditional distribution over the five candidate label strings, whereas the length-normalized variant is a widely used heuristic without that probabilistic interpretation.
To preserve the tokenizer boundary, for each label we form a leading-space continuation and jointly tokenize the canonical prompt-plus-continuation string with \texttt{add\_special\_tokens=False}.
We locate the candidate suffix after the longest stable prefix shared with the standalone prompt tokenization, then verify that decoding the reconstructed prompt and candidate token IDs reproduces the canonical string after trimming surrounding whitespace.
For each candidate, the saved artifacts retain its token IDs, decoded token pieces, token count, and summed and mean log-likelihoods.

\paragraph{Encoder softmax (PubMedBERT).}
The fine-tuned classification head produces logits over five classes, which are softmax-normalized to obtain probabilities.

\subsection{Prompt Templates}

We evaluate four few-shot answer-text prompt templates for the decoder models.
All four templates use the same \emph{verbalizer} (the literal class names \textsc{Background}, \textsc{Objective}, etc.) as continuation targets, so they share one label-to-token mapping.
They differ in instructions, examples, delimiters, and continuation framing:

\begin{itemize}
\item \textbf{Struct-Delim} (\emph{structured-delimiter}): Wraps the input sentence in triple-quote delimiters with explicit category definitions.
  Selected by the AlphaEvolve search.
\item \textbf{Bare-Cont} (\emph{bare-continuation}): Uses bare text without delimiters and the same category definitions.
  Selected by the AlphaEvolve search.
\item \textbf{Min-Inst} (\emph{minimal-instruction}): Concise task instruction with domain-shifted few-shot exemplars (diabetes/GLP-1 domain).
  LLM-designed to test robustness to instruction verbosity and exemplar domain.
\item \textbf{Alt-Frame} (\emph{alternate-frame}): Changes the continuation frame (``Input:''/``Role:'' instead of ``Sentence:''/``Rhetorical Role:'') with distinct exemplars (chronic kidney disease domain).
  LLM-designed to test sensitivity to the lexical frame.
\end{itemize}

All four templates share a common answer-text scoring interface: (1)~a task preamble, (2)~five few-shot demonstrations (one per class), and (3)~a natural continuation point for answer-text log-likelihood scoring.
Struct-Delim and Bare-Cont were selected by the AlphaEvolve search described next, and results using them are exploratory.
Min-Inst and Alt-Frame were designed independently of that search, with different exemplar domains, to probe prompt-template sensitivity.
PubMedBERT does not use prompts (direct text-to-label classification).

\subsection{Evolutionary Prompt Optimization}

A zero-shot baseline prompt collapsed to a single predicted label on a 100-example selection subset, which motivated automated prompt search.
The answer-text scorer used at that stage was preliminary: it could omit the first label token when prompt and candidate tokenization differed at their junction.
We used AlphaEvolve~\citep{alphaevolve2025}, an LLM-driven program-evolution framework, to run a small evolutionary search that started from this baseline prompt.
At each step, AlphaEvolve proposed a complete candidate prompt by expanding or modifying an earlier one, including its instructions, category definitions, continuation frame, and embedded few-shot demonstrations.
We scored each candidate with BioMistral-7B at FP16 on the 100 labeled PubMed RCT examples using our own multi-objective evaluator, and these scores guided selection.
The search evaluated 29 candidates over ${\sim}$3 hours on one H100 GPU.

AlphaEvolve generated the five demonstrations embedded in each selected prompt as part of the candidate text (generic trial sentences, e.g., ``drug X'' versus placebo); the 100 labeled examples served only to score candidates and never appear inside a prompt.
However, those 100 examples came from the test split, and selection relied on the preliminary scorer and a single model.
We later switched the development subset to validation data, but this does not restore held-out status for prompts already selected.
Two template families emerged at the Pareto frontier: Struct-Delim (calibration-optimized) and Bare-Cont (accuracy-optimized).
The selected templates combined few-shot examples, category descriptions, and an answer-text-compatible continuation, features that were not present together in any hand-designed template.

\subsection{Quantization Conditions}

We evaluate three quantization conditions. FP16 serves as the half-precision baseline, requiring approximately 14~GB of weight storage for a 7B model, excluding runtime overhead. INT8 uses LLM.int8() mixed-precision decomposition~\citep{dettmers2022llmint8} via \texttt{bitsandbytes}; outlier features remain in FP16, while the remaining weights are quantized to 8-bit integers, roughly halving weight storage relative to FP16. INT4 uses NF4 (4-bit NormalFloat) quantization with double quantization~\citep{dettmers2023qlora} via \texttt{bitsandbytes}, reducing weight storage by roughly 75\% relative to FP16.

\subsection{Calibration}

We apply post-hoc temperature scaling~\citep{guo2017}, fitting a scalar temperature $T$ on $n = 200$ validation-set examples (seed~42) to minimize NLL, then applying the fitted temperature to $n = 200$ held-out test examples:
\[
P_{\text{cal}}(\ell) = \text{softmax}\!\left(\frac{\text{score}(\ell)}{T}\right)
\]
A fitted $T > 1$ indicates that the induced distribution is more peaked than warranted by accuracy under the selected scorer; $T \approx 1$ indicates the scores are already well-scaled.

\subsection{Metrics}

We evaluate classification performance using accuracy and macro-F1, and calibration using Expected Calibration Error (ECE; 15 equal-width bins), the Brier score, and negative log-likelihood (NLL). Majority-class and class-prior-matched random classifiers provide lower-bound baselines; the latter values are analytical expectations under independent draws from the empirical class distribution.
We computed bootstrap confidence intervals only for the paired accuracy and macro-F1 differences reported in the text.
We report ECE, Brier score, and NLL as point estimates without an uncertainty analysis, so small differences in these metrics should not be read as statistically reliable.

\section{Results}

\subsection{Main Comparison}

Table~\ref{tab:main} presents the main results at $n = 2{,}000$.

\begin{table}[t]
\caption{Main results: sum log-likelihood scoring, uncalibrated, natural-distribution test sample ($n = 2{,}000$).}
\label{tab:main}
\centering
\scriptsize
\setlength{\tabcolsep}{2.5pt}
\begin{tabular}{@{}lllccccc@{}}
\toprule
\textbf{Model} & \textbf{Tmpl} & \textbf{Prec} & \textbf{Acc} & \textbf{F1} & \textbf{ECE} & \textbf{Brier} & \textbf{NLL} \\
\midrule
\multicolumn{2}{l}{PubMedBERT\textsuperscript{*}} & FP32 & .827 & .825 & .072 & .254 & --- \\
\midrule
\multirow{6}{*}{Instruct} & \multirow{3}{*}{S-D} & FP16 & .691 & .624 & .270 & .571 & 2.36 \\
& & INT8 & .698 & .629 & .262 & .558 & 2.26 \\
& & INT4 & .698 & .630 & .261 & .559 & 2.12 \\
\cmidrule(l){2-8}
& \multirow{3}{*}{B-C} & FP16 & .667 & .591 & .211 & .532 & 1.41 \\
& & INT8 & .673 & .604 & .196 & .522 & 1.35 \\
& & INT4 & .660 & .586 & .221 & .551 & 1.53 \\
\midrule
\multirow{6}{*}{BioMistral} & \multirow{3}{*}{S-D} & FP16 & .632 & .589 & .117 & .546 & 1.12 \\
& & INT8 & .636 & .595 & .116 & .539 & 1.10 \\
& & INT4 & .636 & .580 & .087 & .522 & 1.07 \\
\cmidrule(l){2-8}
& \multirow{3}{*}{B-C} & FP16 & .695 & .616 & .104 & .448 & 0.92 \\
& & INT8 & .706 & .631 & .098 & .441 & 0.91 \\
& & INT4 & .700 & .619 & .063 & .440 & 0.90 \\
\midrule
\multirow{6}{*}{Base} & \multirow{3}{*}{S-D} & FP16 & .361 & .276 & .159 & .755 & 1.46 \\
& & INT8 & .403 & .304 & .120 & .722 & 1.41 \\
& & INT4 & .327 & .250 & .217 & .791 & 1.51 \\
\cmidrule(l){2-8}
& \multirow{3}{*}{B-C} & FP16 & .532 & .444 & .063 & .611 & 1.25 \\
& & INT8 & .526 & .439 & .071 & .616 & 1.26 \\
& & INT4 & .565 & .478 & .067 & .581 & 1.22 \\
\midrule
\multicolumn{2}{l}{\textit{Majority}} & --- & .335 & .100 & --- & --- & --- \\
\multicolumn{2}{l}{\textit{Prior-random}} & --- & .261 & .200 & --- & --- & --- \\
\bottomrule
\end{tabular}

\vspace{2pt}
\raggedright\scriptsize S-D = Struct-Delim; B-C = Bare-Cont.
Instruct was evaluated without its chat template.
\textsuperscript{*}PubMedBERT: balanced sample ($n\!=\!2{,}000$, 400/class), ${\sim}176$K supervised examples; contextual reference only, not directly comparable.
\end{table}

The fine-tuned encoder (PubMedBERT) achieves 82.7\% accuracy but, as noted under Table~\ref{tab:main}, it is a contextual reference that is not directly comparable to the few-shot decoder evaluations.

Among decoder models, BioMistral Bare-Cont achieves the highest accuracy (0.706 at INT8), followed closely by Instruct Struct-Delim (0.698 at INT4/INT8).
The generic base model (Mistral-7B-v0.3) performs substantially worse, with Struct-Delim accuracy near the majority baseline (0.327 to 0.403).
Both adapted checkpoints outperform the base model, although the design does not isolate the contribution of either training intervention.
On Bare-Cont, the base model reaches 0.532 to 0.565, which shows strong prompt sensitivity even without adaptation.

Apparent calibration and accuracy diverge between models under sum log-likelihood scoring: BioMistral achieves low ECE (0.063 to 0.117), Instruct has 2 to 3$\times$ higher ECE (0.196 to 0.270), and the base model varies widely by template (0.063 to 0.217).
However, mean-token scoring reverses this gap in all six matched BioMistral-Instruct conditions (Table~\ref{tab:mean_scoring}), so these ECE differences depend on the chosen scorer.

\subsection{The Accuracy-Calibration Tradeoff}

Across six matched BioMistral-Instruct pairs (two templates $\times$ three precisions), Instruct achieves modestly higher mean accuracy ($+$1.4pp) but $2.4{\times}$ higher mean ECE (0.237 vs.\ 0.097) under sum scoring.
The accuracy advantage is template-dependent: Instruct leads on Struct-Delim by $+$5.9 to 6.2pp (paired cluster-bootstrap 95\% CIs exclude zero) but BioMistral leads on Bare-Cont by $+$2.8 to 3.9pp (CIs: $[1.1, 5.5]$pp).
This template-dependent sign change shows that the comparison between these models is not a simple instruction-tuning effect (see Table~\ref{tab:lineage}).

\begin{figure}[t]
\centering
\includegraphics[width=\columnwidth]{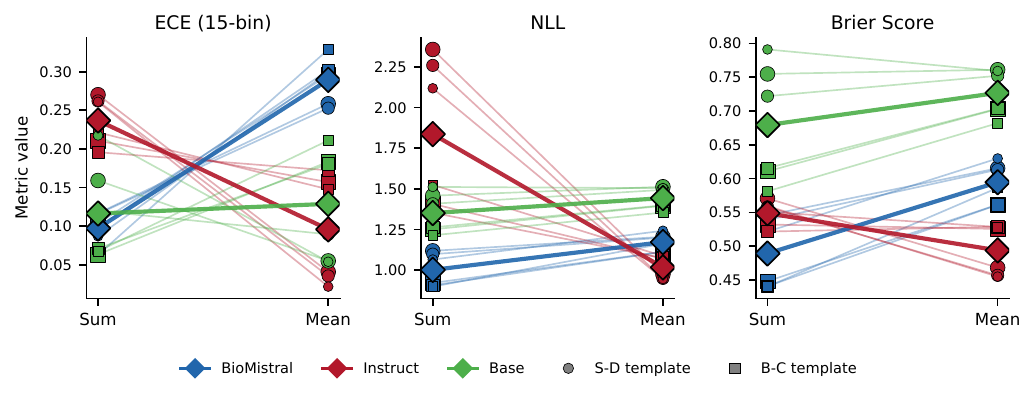}
\caption{Scoring-rule sensitivity across three calibration metrics. Each thin line connects a single model-template-precision condition under summed vs.\ mean-token scoring; diamonds show per-model aggregates. All values are uncalibrated. The BioMistral-Instruct ranking reversal appears in ECE, NLL, and Brier, so it is not specific to ECE binning.}
\label{fig:ece_reversal}
\end{figure}

\paragraph{Scoring-rule reversal.} Table~\ref{tab:mean_scoring} and Figure~\ref{fig:ece_reversal} rescore the same configurations under mean-token log-likelihood.
At FP16, Instruct Struct-Delim ECE drops from 0.270 to 0.041, while BioMistral Struct-Delim ECE rises from 0.117 to 0.259.
Accuracy changes by ${\leq}1.4$pp for the adapted checkpoints but 2.8 to 5.7pp for the base model.
In these configurations, the apparent accuracy-calibration tradeoff therefore depends on the scoring rule.

\begin{table}[t]
\caption{Mean-token log-likelihood scoring, uncalibrated, on the same test sample as Table~\ref{tab:main} ($n = 2{,}000$). The BioMistral-Instruct calibration ranking is reversed relative to sum scoring, while accuracy is largely unchanged for the adapted checkpoints.}
\label{tab:mean_scoring}
\centering
\scriptsize
\setlength{\tabcolsep}{3pt}
\begin{tabular}{@{}llccccc@{}}
\toprule
\textbf{Model} & \textbf{Template} & \textbf{Prec} & \textbf{Acc} & \textbf{ECE} & \textbf{Brier} & \textbf{NLL} \\
\midrule
\multirow{6}{*}{Instruct} & \multirow{3}{*}{Struct-Delim} & FP16 & .694 & .041 & .468 & 0.976 \\
& & INT8 & .701 & .035 & .457 & 0.950 \\
& & INT4 & .697 & .022 & .455 & 0.947 \\
\cmidrule(l){2-7}
& \multirow{3}{*}{Bare-Cont} & FP16 & .664 & .157 & .526 & 1.072 \\
& & INT8 & .668 & .172 & .529 & 1.075 \\
& & INT4 & .659 & .148 & .529 & 1.087 \\
\midrule
\multirow{6}{*}{BioMistral} & \multirow{3}{*}{Struct-Delim} & FP16 & .635 & .259 & .615 & 1.206 \\
& & INT8 & .629 & .253 & .613 & 1.201 \\
& & INT4 & .650 & .299 & .630 & 1.243 \\
\cmidrule(l){2-7}
& \multirow{3}{*}{Bare-Cont} & FP16 & .697 & .296 & .562 & 1.114 \\
& & INT8 & .702 & .302 & .561 & 1.113 \\
& & INT4 & .705 & .329 & .586 & 1.163 \\
\midrule
\multirow{6}{*}{Base} & \multirow{3}{*}{Struct-Delim} & FP16 & .304 & .055 & .761 & 1.513 \\
& & INT8 & .346 & .089 & .752 & 1.494 \\
& & INT4 & .280 & .054 & .759 & 1.504 \\
\cmidrule(l){2-7}
& \multirow{3}{*}{Bare-Cont} & FP16 & .487 & .183 & .703 & 1.396 \\
& & INT8 & .481 & .181 & .704 & 1.399 \\
& & INT4 & .537 & .211 & .682 & 1.355 \\
\bottomrule
\end{tabular}
\end{table}

\subsection{Quantization Stability}

Table~\ref{tab:main} reports FP16, INT8, and INT4 point estimates for each configuration; selected paired bootstrap intervals are reported below.
For BioMistral and Instruct, INT8 accuracy deltas are ${\leq}1.1$ percentage points (the largest being BioMistral Bare-Cont at $+$1.05pp).
Where the 95\% CI excludes zero, the direction uniformly favors INT8 (e.g., BioMistral Bare-Cont: $\Delta\text{F1} = +.015$ $[+.006, +.024]$; Instruct Struct-Delim: $\Delta\text{Acc} = +.007$ $[+.001, +.013]$).

The base model exhibits a larger INT8 effect on Struct-Delim ($+$4.2pp).
INT4 (NF4) shows mixed effects: for BioMistral, the INT4 ECE point estimate is lower than at FP16 (Bare-Cont: 0.063 vs.\ 0.104) with similar accuracy; for Instruct Bare-Cont, INT4 shows a modest accuracy drop ($-$0.7pp) and a slightly higher ECE point estimate.
Because the calibration metrics have no interval estimates, we do not interpret these ECE differences as reliable improvements or degradations.
INT4 does not show the catastrophic degradation sometimes reported for aggressive quantization, though the heterogeneous pattern warrants per-configuration validation before deployment.

\subsection{Temperature Scaling Under Sum Scoring}

Table~\ref{tab:calibration} shows the effect of post-hoc temperature scaling under sum log-likelihood scoring.

\begin{table}[t]
\caption{Temperature scaling under sum log-likelihood scoring, fitted on 200 validation examples and evaluated on 200 separate test examples. ECE is uncalibrated and ECE\textsubscript{cal} is after scaling; both are computed on this 200-example sample, so ECE differs from Table~\ref{tab:main} ($n = 2{,}000$; e.g., BioMistral Struct-Delim FP16: .151 here vs.\ .117). Fitted $T > 1$ indicates that the induced distribution is more peaked than warranted by accuracy.}
\label{tab:calibration}
\centering
\scriptsize
\setlength{\tabcolsep}{3pt}
\begin{tabular}{@{}lllrrrl@{}}
\toprule
\textbf{Model} & \textbf{Template} & \textbf{Prec} & \textbf{ECE} & \textbf{ECE\textsubscript{cal}} & $T$ & \textbf{Effect} \\
\midrule
BioMistral & Struct-Delim & FP16 & .151 & .079 & 1.67 & $1.9{\times}\downarrow$ \\
BioMistral & Struct-Delim & INT8 & .188 & .105 & 1.67 & $1.8{\times}\downarrow$ \\
BioMistral & Struct-Delim & INT4 & .099 & .088 & 1.44 & $1.1{\times}\downarrow$ \\
BioMistral & Bare-Cont & FP16 & .153 & .072 & 1.44 & $2.1{\times}\downarrow$ \\
BioMistral & Bare-Cont & INT8 & .142 & .061 & 1.44 & $2.3{\times}\downarrow$ \\
BioMistral & Bare-Cont & INT4 & .105 & .085 & 1.25 & $1.2{\times}\downarrow$ \\
\midrule
Instruct & Struct-Delim & FP16 & .307 & .073 & 4.64 & $4.2{\times}\downarrow$ \\
Instruct & Struct-Delim & INT8 & .295 & .083 & 4.01 & $3.5{\times}\downarrow$ \\
Instruct & Struct-Delim & INT4 & .309 & .081 & 4.01 & $3.8{\times}\downarrow$ \\
Instruct & Bare-Cont & FP16 & .261 & .066 & 2.59 & $3.9{\times}\downarrow$ \\
Instruct & Bare-Cont & INT8 & .243 & .082 & 2.23 & $3.0{\times}\downarrow$ \\
Instruct & Bare-Cont & INT4 & .266 & .098 & 2.59 & $2.7{\times}\downarrow$ \\
\bottomrule
\end{tabular}
\end{table}

ECE point estimates decrease for both models: BioMistral fitted temperatures ($T = 1.25$ to $1.67$) correspond to modest ECE reductions ($1.1$ to $2.3{\times}$); Instruct temperatures are higher ($T = 2.2$ to $4.6$), with $2.7$ to $4.2{\times}$ reductions.
These ratios are computed from point estimates on 200 examples without confidence intervals, so the smaller reductions (e.g., $1.1{\times}$) should not be read as reliable improvements.
This result is specific to sum scoring; we did not fit temperatures under mean-token scoring, where the calibration ranking reverses (Table~\ref{tab:mean_scoring}).
Fitted temperatures vary modestly across precisions (e.g., Instruct Struct-Delim $T = 4.64$ at FP16 vs.\ $4.01$ at INT8/INT4), though the 64-point grid limits resolution.

\subsection{Template Sensitivity}

We evaluate Struct-Delim and Bare-Cont on the test set ($n = 2{,}000$) and Min-Inst and Alt-Frame on a validation set ($n = 200$).
Because the two pairs use different samples, we compare template effects only within each pair.

\begin{table}[t]
\caption{Auxiliary-template sensitivity under sum log-likelihood scoring, uncalibrated: per-model averages across FP16, INT8, and INT4 for Min-Inst and Alt-Frame on the validation-derived sample ($n = 200$). Mean-token results are not reported for these templates.}
\label{tab:template_sensitivity}
\centering
\scriptsize
\setlength{\tabcolsep}{3pt}
\begin{tabular}{@{}llcccc@{}}
\toprule
\textbf{Model} & \textbf{Tmpl} & \textbf{Acc} & \textbf{F1} & \textbf{ECE} & \textbf{Brier} \\
\midrule
\multirow{2}{*}{Instruct} & Min-Inst & .695 & .636 & .171 & .485 \\
& Alt-Frame & .717 & .641 & .207 & .461 \\
\midrule
\multirow{2}{*}{BioMistral} & Min-Inst & .592 & .515 & .110 & .551 \\
& Alt-Frame & .570 & .485 & .081 & .610 \\
\midrule
\multirow{2}{*}{Base} & Min-Inst & .382 & .346 & .177 & .740 \\
& Alt-Frame & .455 & .390 & .132 & .684 \\
\bottomrule
\end{tabular}
\end{table}

Table~\ref{tab:template_sensitivity} shows validation-set accuracy differences of 2.2pp for Instruct, 2.2pp for BioMistral, and 7.3pp for the base model between the two auxiliary templates.
Across the two primary templates in Table~\ref{tab:main}, the corresponding per-precision differences are 2.4 to 3.8pp for Instruct, 6.3 to 7.0pp for BioMistral, and 12.3 to 23.8pp for the base model.
Among the auxiliary templates, BioMistral's accuracy and calibration estimates favor different templates (Min-Inst: higher accuracy; Alt-Frame: lower ECE), while Instruct's higher-accuracy template (Alt-Frame, 0.717) also has the higher ECE (0.207).
At $n = 200$ and without interval estimates, these ECE differences may not be distinguishable from sampling variation.
Together with the scoring-rule result above, these observations show that apparent calibration varied with both template and scorer in this study.
We report Min-Inst and Alt-Frame under summed scoring only, so they show prompt sensitivity and do not independently confirm the scoring-rule reversal.

\subsection{Per-Class Performance}

Table~\ref{tab:perclass} shows per-class F1 at FP16 under sum scoring.
Performance varies sharply across classes: all models achieve their highest F1 on \textsc{Methods} (0.56 to 0.82) and lowest on \textsc{Conclusions} (0.01 to 0.46).
The \textsc{Conclusions} class has the longest label name (5 tokens), consistent with a surface-form length penalty under sum scoring.
The base model nearly collapses \textsc{Conclusions} on Struct-Delim (F1 = 0.013) while maintaining moderate \textsc{Methods} performance.

\begin{table}[t]
\caption{Per-class F1 at FP16 under sum log-likelihood scoring, uncalibrated, on the test sample ($n = 2{,}000$). \textsc{Methods} is the easiest class across all models; \textsc{Conclusions} (longest label, 5 tokens) is hardest.}
\label{tab:perclass}
\centering
\scriptsize
\setlength{\tabcolsep}{3pt}
\begin{tabular}{@{}llccccc@{}}
\toprule
\textbf{Model} & \textbf{Tmpl} & \textbf{BG} & \textbf{OBJ} & \textbf{MTH} & \textbf{RES} & \textbf{CON} \\
\midrule
Instruct & Struct-Delim & .472 & .603 & .819 & .761 & .464 \\
Instruct & Bare-Cont & .474 & .598 & .818 & .707 & .360 \\
BioMistral & Struct-Delim & .526 & .599 & .710 & .675 & .432 \\
BioMistral & Bare-Cont & .587 & .582 & .811 & .728 & .372 \\
Base & Struct-Delim & .268 & .104 & .564 & .430 & .013 \\
Base & Bare-Cont & .350 & .476 & .775 & .477 & .140 \\
\bottomrule
\end{tabular}
\end{table}

\subsection{Failure of Single-Token Class-Code Scoring}

Before adopting answer-text scoring, we evaluated four class-code templates in nine interpretable BioMistral template-precision configurations ($n = 2{,}000$ each) using restricted softmax over single-letter logits (A to E): FP16 and INT8 for all four templates, plus INT4 for one template.
We excluded an additional label-permutation condition because its outputs could not be interpreted under the intended mapping.
Among the nine retained configurations, five predicted a single class for all examples, and macro-F1 ranged from 0.038 to 0.095.
These results document severe collapse under the evaluated class-code prompts but do not identify whether prompt design, code-position bias, or another factor caused it.
This failure motivated the full-label answer-text protocol used in the main experiments; because the prompts and scoring protocol changed together, the preliminary and main results are not a controlled comparison of scoring methods alone.

\section{Discussion}

\subsection{Scoring Protocol and Calibration}

Within this exploratory design, our central observation is that, for the evaluated configurations, templates, and samples, the calibration ranking of BioMistral and Instruct, measured without post-hoc calibration, reverses between summed and mean-token log-likelihood.
Under summed scoring, Instruct appears less well calibrated than BioMistral (average ECE 0.237 vs.\ 0.097); under mean-token scoring, the order flips.

Temperature is an unresolved alternative explanation for the reversal.
Mean-token normalization divides each candidate's summed log-likelihood by its token count $T_\ell$, which compresses score differences before the softmax.
If all labels had the same length $T$, this would equal temperature scaling of the summed scores with temperature $T$; with label lengths of 3 to 5 tokens the correspondence is only approximate, because a label-specific divisor can also reorder candidates.
The reversal may therefore partly reflect a difference in probability sharpness between the two scorers.
The temperatures fitted under summed scoring (Instruct $2.2$ to $4.6$, near the label-length range; BioMistral $1.3$ to $1.7$, below it) are compatible with this account but, being fitted on 200 examples under one scorer, do not test it.
We did not fit a temperature for each scorer, so we cannot say whether the reversal would persist, shrink, or disappear after per-scorer calibration.

This observation has a methodological implication: a calibration comparison made under a single scoring protocol may not hold under an alternative protocol that is equally common in practice.
We do not claim that calibration is intrinsically or universally unstable: the primary templates were selected using test-derived data, and the reversal was not tested after per-scorer calibration.

\subsection{Sources of Variation}

Our results suggest a qualitative ordering of effect sizes on this task, although the ranking is itself sensitive to the scoring rule. Among the factors we varied, the scoring rule produced the largest change in apparent calibration: switching from sum to mean log-likelihood reverses the BioMistral-Instruct ranking while barely affecting their accuracy. We include PubMedBERT only as a contextual reference; its supervision, architecture, and evaluation sample all differ from the decoder runs, so we do not attribute its higher accuracy to any single factor.

Across the two primary templates, template choice changes mean accuracy across precisions by ${\sim}$3pp for Instruct, ${\sim}$7pp for BioMistral, and up to ${\sim}$18pp for Base. The Instruct-vs-BioMistral accuracy difference is smaller ($+$1.4pp on average under sum scoring) and changes sign on Bare-Cont. Quantization generally has the smallest effect for the adapted checkpoints (INT8: ${\leq}$1.1pp), whereas base-model accuracy shifts by up to $+$4.2pp at INT8 and by approximately $-$3.5pp to $+$3.3pp at INT4, depending on the template.

\subsection{Implications for Evaluation Practice}

For biomedical NLP evaluation, calibration comparisons between decoder models should report results under multiple candidate-scoring normalizations (at minimum, summed and mean-token log-likelihood) and identify which conclusions are robust to that choice. Evaluations should likewise treat template design as an experimental variable, not an implementation detail, given the within-pair accuracy swings observed here. INT8 quantization produces small accuracy changes (${\leq}$1.1pp) for the adapted checkpoints studied, although the base model shows larger shifts; therefore, equivalence testing with predeclared margins is needed before claiming deployment safety.

\subsection{Limitations and Future Work}

Our evaluation is limited to a single task, PubMed RCT. Generalization to other biomedical classification tasks (e.g., relation extraction and clinical NLI) remains to be validated.
The range of compression methods is also limited: we evaluate only \texttt{bitsandbytes}-based INT8 and NF4 (INT4) quantization. Other approaches, including GPTQ~\citep{frantar2023gptq}, AWQ~\citep{lin2024awq}, and knowledge distillation, may exhibit different calibration-accuracy tradeoffs.

The results are sensitive to the verbalizer and scoring rule. Answer-text scoring uses label names of unequal token length (e.g., RESULTS = 3 tokens, CONCLUSIONS = 5), creating a surface-form length prior.
The low \textsc{Conclusions} F1 under sum scoring (Table~\ref{tab:perclass}) is consistent with, but does not by itself establish, a surface-form length penalty.
Because raw and length-normalized likelihoods can both remain choice-sensitive, robust evaluation should compare multiple scoring normalizations~\citep{cho2026choices}.
Equal-length verbalizers and label-permutation controls are additional checks for future work.

Prompt selection introduces a provenance caveat: we selected Struct-Delim and Bare-Cont in July 2026 on examples from the PubMed RCT \emph{test} split with a flawed preliminary scorer (see Methods) and did not re-select them afterwards.
Results using these templates are therefore exploratory rather than the product of a clean held-out selection procedure.

The temperature-scaling analysis is further constrained by its sample size (200 validation examples for fitting and 200 test examples for evaluation), which yields noisier ECE estimates and fitted temperatures than the $n = 2{,}000$ main evaluation.
The fitted temperatures come from a 64-point geometric grid over $[0.01, 100]$ and reflect that resolution rather than continuous optima.

Finally, the model comparisons are confounded: BioMistral and Instruct differ in more than one training variable (Table~\ref{tab:lineage}), Instruct was evaluated without its chat template, which may independently affect calibration~\citep{sanzguerrero2026overconfident}, and PubMedBERT differs from the decoders in supervision and evaluation sample.
We evaluate 7B-parameter decoders only; larger models may exhibit different calibration dynamics.

\section{Conclusion}

Motivated by resource-constrained local deployment, we evaluated confidence reliability for quantized biomedical language models on public PubMed abstracts. Within an exploratory design affected by test-derived prompt selection and a token-boundary error in the preliminary selection scorer, we observed that scoring normalization reversed the apparent calibration ranking of BioMistral and Instruct, measured without post-hoc calibration, while leaving accuracy largely unchanged; we did not test whether this reversal persists after per-scorer calibration. We also observed that prompt-template choice can substantially alter performance. Quantization effects are model- and template-dependent, with lower-bit quantization producing less consistent behavior. These observations motivate treating prompt design and scoring normalization as first-order experimental decisions and validating compression choices for each deployment setting in clean held-out studies.

\bibliography{references}

\end{document}